\documentclass[journal]{IEEEtran}

\ifCLASSINFOpdf
\else
\fi

\usepackage{graphicx}
\usepackage{amsmath}
\usepackage{array}
\usepackage{caption}
\usepackage{booktabs}
\usepackage{tabularx}
\usepackage{multirow}
\usepackage{titlesec} 
\usepackage{hyperref}
\usepackage[font=small,labelfont=small,labelsep=period]{caption}
\usepackage{cite}
\usepackage{placeins}
\usepackage{orcidlink}
\usepackage{float}
\usepackage{gensymb}

\begin{document}

\title{Performance of Human Annotators in Object Detection and Segmentation of Remotely Sensed Data}

\author{
    Roni~Blushtein-Livnon\raisebox{0.5ex}{\orcidlink{0000-0002-3493-4894}}, 
    Tal~Svoray\raisebox{0.5ex}{\orcidlink{0000-0003-2243-8532}}, 
    and~Michael Dorman\raisebox{0.5ex}{\orcidlink{0000-0001-6450-8047}}
    \thanks{R. Blushtein-Livnon and M. Dorman are with the Department of Environmental, Geoinformatics and Urban Planning Sciences, Ben-Gurion University of the Negev, Israel (e-mail: livnon@bgu.ac.il; dorman@bgu.ac.il).}
    \thanks{T. Svoray is with the Department of Environmental, Geoinformatics and Urban Planning Sciences, and the Department of Psychology, Ben-Gurion University of the Negev, Israel (e-mail: tsvoray@bgu.ac.il).}%
}

\maketitle

\begin{abstract}
This study introduces a laboratory experiment designed to assess the influence of annotation strategies, levels of imbalanced data, and prior experience, on the performance of human annotators. The experiment focuses on labeling aerial imagery, using ArcGIS Pro, to detect and segment small-scale PVs, selected as a case study for rectangular objects. The experiment is conducted using images with a pixel size of 0.15\textbf{$m$}, involving both expert and non-expert participants, across different setup strategies and target-background ratio datasets. Our findings indicate that annotators generally perform more effectively in object detection than in segmentation tasks. A marked tendency to commit more Type II errors (False Negatives, i.e., undetected objects) than Type I errors (False Positives, i.e. falsely detecting objects that do not exist) was observed across all experimental setups and conditions, suggesting a consistent bias in annotation processes. Performance was better in tasks with higher target-background ratios (i.e., more objects per unit area). Prior experience did not significantly impact performance and may, in some cases, even lead to overestimation in segmentation. These results provide evidence that annotators are relatively cautious and tend to identify objects only when they are confident about them, prioritizing underestimation over overestimation. Annotators' performance is also influenced by object scarcity, showing a decline in areas with extremely imbalanced datasets and a low ratio of target-to-background. These findings may enhance annotation strategies for remote sensing research while efficient human annotators are crucial in an era characterized by growing demands for high-quality training data to improve segmentation and detection models.  
\end{abstract}


\IEEEpeerreviewmaketitle

\section{Introduction}

\IEEEPARstart{A}{vailability} of reliable training data is a prime requirement for computer vision tasks such as automatic object detection (OD) and segmentation. Even large generic models, such as the \textit{Segment Anything Model (SAM)}, that promise zero-shot capabilities, require problem-specific data to adapt to the particular problem being studied, often through methods such as transfer learning \cite{10378323}. Specifically, training sets are much needed for many remote sensing (RS) applications for earth observation missions \cite{chen2024changedetectionopticalremote}. Despite recent developments in automatic, or semi-automatic, architectures to create training datasets \cite{8099989} \cite{huang2024data}, and the availability of historical datasets for reuse, human annotators are still identified as the major source for the generation of training sets for computer vision tasks \cite{8237359} \cite{10378323} \cite{lin2014microsoft}. This is also true in other artificial intelligence fields such as natural language processing (NLP) \cite{stenetorp2012brat}. The need for data annotation (or labeling) by humans is particularly actual to measure performance of classifiers on quantitative tasks such as density measurement of rooftops. As machine learning (ML) usage increases dramatically \cite{sharifani2023machine} for various purposes, so is the need for human-annotated gold standards \cite{aroyo2015truth}. This requirement increases due to the emerging usage of crowdsourcing procedures and the ongoing development of large training sets by multiple nonexpert annotators \cite{lu2020research}, which increase motivations to advance knowledge on how to increase labeling efficiency. 

Human annotators vary in their skills, capabilities, and the approaches they use to achieve optimal annotation. Some annotators may have years of relevant experience while others may be laymen. Moreover, the approach in which the annotation process is conducted may also affect the outcome quality, for instance, whether the final annotation is based on a group of annotators and how each annotator influences the final result. Annotators can also be influenced by the object, or the task, characteristics. In RS applications, annotation by humans can be particularly challenging for various reasons, such as objects within the same category appearing differently in different images or regions within the same image. For example, a solar panel, which may be assumed a simple rectangular object, can be oriented in multiple directions toward the sun, affected by internal and external shading, and can vary in color and size. Consequently, it is difficult to expect that two random human annotators will produce the same results. Such differences may have substantial implications for training and prediction processes. Previous Earth observation studies have demonstrated that variations in training sets can impact training and validation accuracy. For instance, by using synthetic golden standards through a simulation process of real ground data, \cite{geomatics4010005} and \cite{10.1371/journal.pone.0291908} demonstrate that the direction and magnitude of accuracy metric mis-estimation were a function of prevalence, size, and nature of imperfections in the reference standard. Namely, ground data that introduced errors and biases led to an incorrect underestimation of the model's performance. While studies based on synthetic models have significant implications for understanding ground data accuracy, they are limited to simulated human behavior. This means they do not capture the actual biases, limitations, and advantages of human actions but instead generalize them. Findings based on simulations can reflect generic patterns but do not provide particular evidence of actual human actions and behaviors. Understanding the latter is crucial and requires further exploration, as will be demonstrated below.  

Our \textit{aim} is to assess performance of human annotators in two key computer-vision tasks: object detection and segmentation of RS data. Specifically, we tackle the following four \textit{operative objectives}:
(1)	To compare the performance of human annotators in OD versus segmentation tasks;
(2)	To analyze distinctions between Type I errors (False Positives) and Type II errors (False Negatives) in annotations across different task conditions (sparse-target task versus dense-target task) and annotation setup strategies (individual versus group; and independent versus dependent group setup processes);
(3)	To compare annotators' performances under various task conditions and setup strategies.
(4)	To examine the impact of prior experience in RS data interpretation, digitization, and annotation on the annotation performance.

To achieve these objectives, we conducted an experiment on photovoltaic solar panels (PV), as a case study for rectangular objects with varied appearances, in a semiarid environment. The experiment involved 24 human participants and was carried out in a geoinformatics laboratory. The findings enhance the development of training sets by humans and improve our understanding of how annotators operate, ultimately leading to better selection and guidance of human annotators.

\section{Related work}
\subsection{Training Sets and Human Annotators in RS}
Image segmentation is broadly referred to the technique of partitioning an image into segments labeled based on their characteristics. Pixel-level labeling, known as semantic segmentation, is an essential computer vision and RS technique for, e.g., crop cover analysis\cite{Jadhav_2018}, land-use mapping \cite{8447270}, and environmental monitoring\cite{yuan2010automatic}. Among various semantic segmentation methodologies, Machine Learning has gained increasing usage over the past decades, with a notable rise in using Deep Learning in recent years \cite{MO2022626}. Consequently, labeling has gained interest by many researchers \cite{YUAN2021114417}. Labeling processes rely on training sets that are accurately labeled using trustworthy sources and then used to train models for predictions. The quality of training sets is crucial, as they influence the performance of machine-  and deep-learning models. Thus, improving training processes to extract high-quality training sets is much sought after. While efforts were made to generate simulated training sets using machines or to enhance human annotation through automated algorithms \cite{xia2015accurate}, most current image segmentation models still rely on human input  (e.g., \cite{lin2014microsoft} \cite{10378323}). 

Annotations by humans can be utilized at the beginning of the training process, creating a high-quality dataset that serves as a reference for model's training and later being expanded automatically \cite{10378323}; or, at the final stage for monitoring objects/pixels predicted by models with low confidence and high loss. For example, \cite{Benenson_2019_CVPR} evaluated annotations quality using data from human annotators who were asked to correct model segmentation with corrective clicks.  

Annotators can be students, professionals, and even random individuals participating in crowdsourcing \cite{9251151}, whereas the latter especially are characterized by varying quality \cite{zhang2023learning} \cite{hobley2023crowdsourcing} and sometimes may even reduce overall performance. The latters are usually referred to as \textit{malicious} workers \cite{qiu2018crowdeval}. Human annotators are among the most commonly used sources in RS data \cite{9035660}, which often lacks sufficient labeled training sources\cite{s20061594} \cite{wang2023accurate}. The demand for reliable data from humans has led to an ongoing effort to develop semi-automated processes for generating RS annotations \cite{BENATO2021107612} \cite{s20061594} \cite{rs13112064} \cite{8898111}. For example, \cite{hua2021semantic} proposed a framework for annotating RS images using humans for initial annotation only, which then serves as a basis for automatic annotation and reduces the efforts required for full annotation. Namely, even in generated annotations, reliance on human annotations at the initial stage is still unavoidable and will probably be inevitable in the near future, while human annotators' quality clearly impacts machine-generated training sets \cite{wang2023accurate}. Human annotation performance can suffer biases and we consider here three of them. The \textit{first} is the annotation strategy, namely how the final annotation is obtained. \textit{Second} is the nature of the task, in particular targets prevalence. The \textit{third} bias is the diversity of human annotators' characteristics, e.g., their prior experience.  

Previous studies did not examine the impact of these biases on human annotation quality and such an investigation would improve understanding of training set generation and lead to more informed recommendations on how to best activate human annotators in creating training sets. This is critical for RS data analysis, because it is more challenging to annotate RS data than simple ground photos due to their complexity in interpretation, richness, diversity, and intricacy of the information they contain, and the vast presence of objects with similar appearances \cite{wang2023accurate}. 

Human annotators quality was investigated, primarily comparing cognitive abilities \cite{IVASICKOS20159539}, demographic characteristics \cite{gruhn2008age}, tactics of annotation process \cite{CHEN2018985}, and their reliability \cite{qiu2018crowdeval} \cite{bragg2016optimal} \cite{northcutt2021pervasive} \cite{lu2020research}. However, these studies have focused on human annotation of standard photographs rather than RS imagery. Only a few studies addressed annotation quality of the latter. Among these, \cite{wang2024well} has found that more educated annotators performed better at interpreting land-cover types from RS imagery. Manual interpretation accuracy improved with training, access to ultra-high-resolution images as supplementary data, and the annotators' familiarity with the study area. The performance of a CNN model, when trained on a small sample dataset, was inferior to that of manual interpretation. Group consistency also proved a reliable indicator of the samples' quality. However, Wang et al. indeed assessed the accuracy and strategy of individual annotators but  not of groups and also did not address the common imbalanced data problem, which is typical to RS tasks \cite{rs15071768}. These two issues, among others, are addressed here.  

\subsection{Annotation Strategy}
Annotating RS data is time-consuming and, therefore, labor-intensive, expensive, and can be error-prone \cite{rs71115014} \cite{YUAN2021114417}. It is, therefore, crucial to select and plan the most useful annotation strategy to ensure the highest performance. Among those available to the researcher are the \textit{individual} and \textit{group} strategies. The individual strategy is the simple activation of a single annotator who is requested to identify an object and delineate its borders according to her own consideration. Such an approach does not mix the knowledge of several individuals and is highly dependent on each annotator's skills and expertise. In the group approach, a team of annotators combine knowledge and detect or segment objects. This is typically accomplished using a majority vote decision \cite{raykar2010learning}\cite{9035660} \cite{10.1162/tacl_a_00449}, where, in binary tasks, individual votes are compared, and the object (in classification tasks) or pixel (in segmentation task) is included in the final annotation if the majority of annotators have marked it. In multi-class tasks, the most frequent label among the annotations (i.e., the majority) is selected as the final label for that object/pixel. This approach was used mainly when non-experts made the pool of annotators and the group compensated for errors \cite{zhang2023learning}. Another group strategy is to assign a greater weight to annotators with higher skills or prior experience. For example, \cite{wang2024well} proposed an annotation framework featuring a pre-annotation test of the annotators to estimate their ability to distinguish between classes. The framework assesses each annotator's labeling proficiency, providing a prior label quality estimate, whereas a subsequent quality evaluation process uses this prior quality as a weight, giving higher weights to annotators who perform better in difficult classes. This approach may increase likelihood for correct labels to prevail, even when they are fewer in number. 
Another group strategy is the sequential approach, where the annotation process is performed in progressive stages: the first annotator labels or delineates the objects, and subsequent annotators review the process one after the other \cite{9251151}.

\subsection{Task Conditions - Level of Imbalanced Data}
High-class imbalance is naturally inherent in many real-world applications, e.g., medical diagnosis, natural disaster prediction, fraud detection, etc. Imbalanced data problems are also very frequent in RS problems, especially when detecting specific objects \cite{YUAN2021114417}, and in both cases in which only two classes are present in the considered data set and in multiclass cases \cite{BRUZZONE19971323}. In deep learning applications in RS, this issue was extensively studied e.g., \cite{buda2018systematic} \cite{johnson2020effects} \cite{valova2020optimization} \cite{9335495}, since highly imbalanced data poses added difficulty, as most models exhibit bias towards the majority class while neglecting underrepresented ones, therefore leading to biased predictions and increased Type I errors (false positives). Imbalance can also slow down learning processes, as the model struggles to learn effectively from limited examples of minority classes and abundance of examples from the majority group, ultimately reducing overall accuracy and generalizability \cite{Johnson_2019} \cite{Bauder_2018} \cite{Ghosh_2022}.   
A key question in human annotators' performance is how imbalanced data affects their ability to identify objects and the tendency to over- or under-estimate their prevalence. In an imbalanced dataset, class distribution is not uniform, and in a binary problem e.g., detection and segmentation of PV cells, the majority of data belongs to one class (non-PV cell), and only a small part belongs to the other (PV cells), a problem often referred to as a needle in a haystack. Thus, the study of large regions with a sparse appearance of a specific target can lead to a decrease in the annotator's performance by decreasing the opportunity to identify the object by increasing the search time of large, sometimes monotone areas that may be tiring to search for a specific object, and by increasing the likelihood of labeling targets where they are not present (false positive). Furthermore, in dense areas, various phases of the same entities can be viewed by human annotators in a small and condensed area, the annotator can become more familiar with the object, exposed to similar targets in proximity, that is, within the same field of view, allowing for comparison between targets, and therefore more easily identify them. Human annotators can use the surrounding context to make informed decisions about ambiguous or partially obscured objects. For example, in an urban environment, the annotator can identify a solar panel when it is located on the roof center because she saw it in many other cases in a similar spot or with the same deployment pattern. In dense areas, this ability may help accurately annotate objects that may be closely located together. Despite their presumed potential effect described above, these biases of the target-background ratio of an RS imagery on human annotation quality have not been investigated so far. 

\subsection{Annotators Expertise}
The ability to digitize and label objects correctly in RS data may differ between individuals, as with any other tasks, and especially complex tasks, that humans apply \cite{xia2015accurate}. Differences in capabilities may be related to spatial cognition of the annotator and her ability to process information about different environments and spaces \cite{barbara}. Spatial cognition is influenced by various personal characteristics that are relevant to annotation tasks, such as wayfinding ability, life stage, gender, and prior experience \cite{Gifford}. Individuals are also prone to spatial distortions, including alignment and rotation biases, distance and direction errors, and structural and semantic biases \cite{tversky2005visuospatial}. While it is beyond the scope of this paper to explore the various aspects of spatial cognition that affect human annotators, we focused on experience and, more specifically, on the difference between experts and non-experts in their performance and their tendency to make mistakes.
This category was chosen because human annotation skills have gained significant attention in the growing field of crowdsourcing. Previous studies indicate that agencies often employ non-experts, who are available at a lower cost, reserving experts for tasks that are particularly complex and justify higher annotation expenses. Segmentation annotations of medical images, for example, suffer from a scarcity of expert annotators, and annotations in that field show considerable variability, which is further influenced by the annotator's level of expertise \cite{kats2019soft} \cite{zhang2023learning}. This raises a key question for annotation projects: which type of annotator is cost-effective to employ? Additionally, the differences between these two groups affect how crowdsourced data should be managed (see the section above on annotation strategy). A critical question arises regarding how this practice impacts the quality of training datasets created by non-experts. This conundrum was also observed in NLP annotations \cite{10534765}.
For example, the study by \cite{aroyo2015truth} demonstrated that contrary to the common belief that "experts are better", experts did not consistently produced higher-quality annotations compared with non-experts. While lay users might make mistakes due to a lack of knowledge, experts possessing sufficient knowledge may sometimes misinterpret implicit data, a tendency not observed in lay users. Conversely, research in RS indicates that generating high-quality labels for RS data is a complex task that demands  previous knowledge \cite{wang2023accurate}. Additionally, \cite{hua2021semantic} found that expert annotators are often necessary for accurately identifying pixels at object boundaries and in ambiguous regions within urban landscapes using RS data. Similar findings were observed in studies examining the impact of expertise on human annotation of medical images, and strategies for properly weighing the annotator's level of expertise have been developed in this field\cite{Warfield2004903} \cite{zhang2023learning}. The work of \cite{wang2024well} showed that annotation accuracy increased after individual annotators were trained and provided with supporting and diverse background material on the problem. However, no study has examined the difference in annotation quality between experts and non-experts in RS imagery and the weight that should be given to these experts in determining the final results when working with a group of annotators using the commonly-used majority vote strategy.

The above four subsections describe distinct challenges of data annotation by humans, and the large need for high-quality training sets in various RS fields of image segmentation. The current paper aims  to bridge the three research gaps and suggests an experiment that provides knowledge on the difference in performance of human annotators, focusing on comparisons between experimental setups, task conditions, and variations in annotators' expertise. According to our best knowledge, such an experiment has not been published yet.

\section{Methods}
\subsection{Participants}
Overall 24 students from the Department of Environmental, Geoinformatics and Urban Planning Sciences at Ben-Gurion University of the Negev, have participated in the experiment. This sample size aligns with commonly accepted standards of human annotators' experiments in RS works \cite{kim2024understanding} \cite{hobley2023crowdsourcing}. All students are graduates of the course 'Introduction to Geoinformatics', during which they have gained limited experience in using various types of RS imagery. The course includes a two-hour class how to perform digitization and further usage of ArcGIS Pro digitizing tools for a single home exercise. So the majority of the participants have a very limited experience that can be acquired by guided learning of a few hours in every agency. Prior to the experiment, the participants were asked to complete a questionnaire detailing their previous experience in interpreting RS imagery, conducting digitization processes, and performing annotation tasks. Among the 24 participants, six experts were identified, with an average prior experience of 22 months in professional annotation projects in the public or private sector. These six participants are denoted hereafter as the experts. The participants' ages ranged from 20 to 30, with 56\% being male and 44\% female. All participants were first- or second-year undergraduate students. 

\subsection{Experimental Setup}
\begin{figure}[h]
    \centering
    \includegraphics[width=1\linewidth]{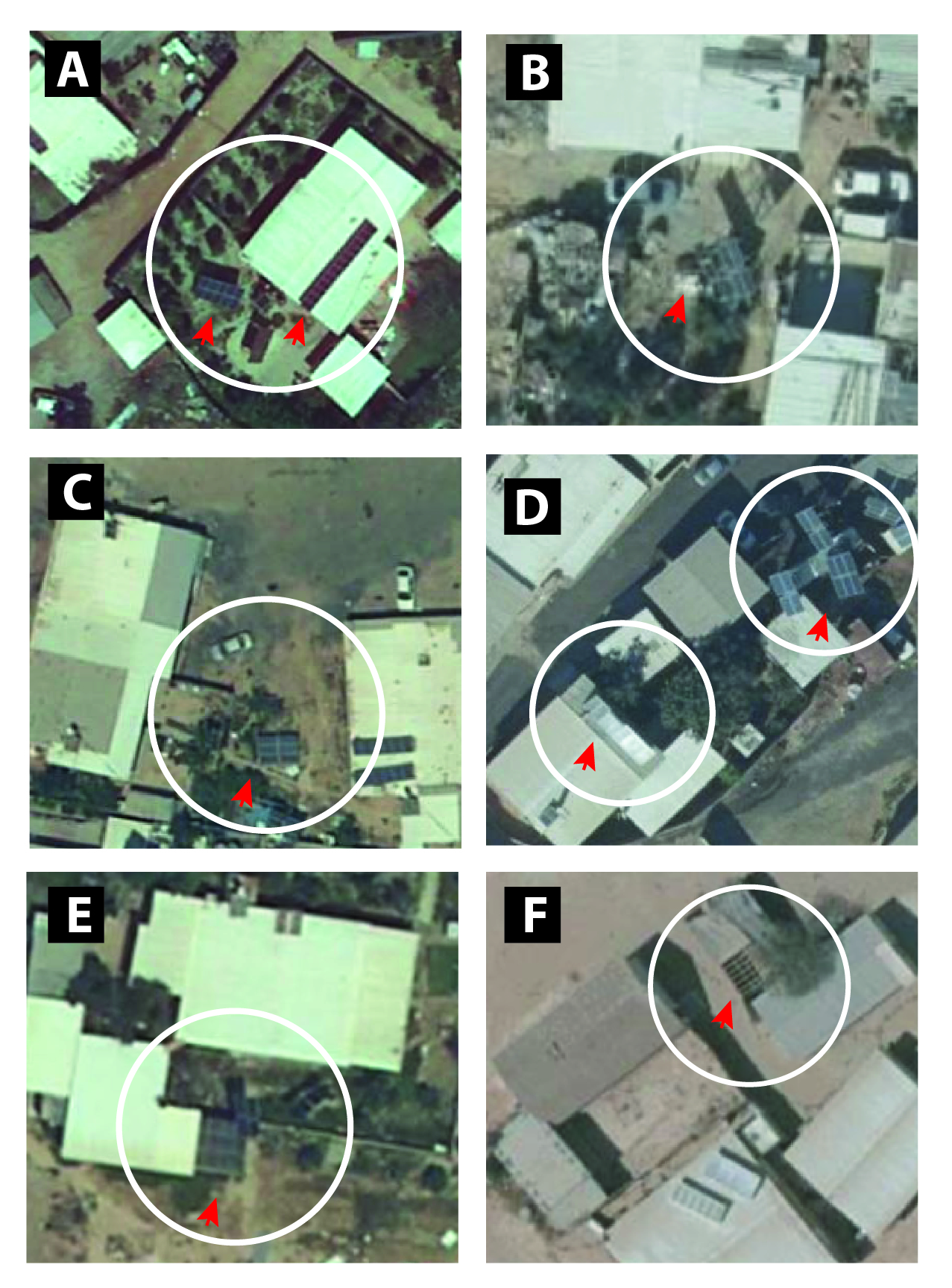}
    \caption{Challenges of small-scale PVs detection from RS images: A - Low contrast of ground-based panels with their surroundings (on the left), compared with a high contrast of rooftop panels (on the right). B - Presence of adjacent objects near ground panels make them difficult to detect. C - Shading of the panel makes it difficult to distinguish it from the target. D - The target appears in varying RGB values, making it difficult to identify. E – resemblance to other objects: a small shade structure with a similar size and color to a solar panel. F - a striped tarpaulin sheet resembling a solar panel.}
    \label{challenges}
\end{figure}

The participants were asked to identify all solar panels appearing in the aerial image (OD task) and to delineate the solar panel boundaries with maximum accuracy (segmentation task). In addition, they were asked to rate each identified object according to three levels of confidence in the correctness of their identification. Before the experiment, they received a detailed briefing on the nature of the target (small-scale solar panels), its common and less common appearances, and the challenges involved in its detection and delineation (such as shading, low contrast, resembling objects, and adjacent objects, see Figure \ref{challenges}). 

\subsubsection{Annotation strategy}
The participants were divided into two equal size groups: one group performed an \textit{independent annotation process}, while the other followed a \textit{dependent annotation process}. The participants in the independent process were asked to perform the annotation tasks independently of other annotators. These participants were randomly assigned to teams of three annotators. The final annotation for each team was determined based on the majority vote principle: for the evaluation of OD performance, an object (panel) labeled by two or more annotators was included in the final annotation. For segmentation evaluation, a pixel marked by two or more annotators was included in the final annotation. The participants in the dependent annotation process were divided into teams of three annotators, where the first conducted a full annotation, the second reviewed the detection and segmentation, and the third reviewed the detection and segmentation after the previous reviewer (see Figure \ref{setup}A). A final annotation was obtained after the second review. Further comparison was made between the annotation performance of individuals, where their annotation served as the final annotation for performance evaluation, and groups, consisting of the final annotation of both dependent and independent teams.
\subsubsection{Task Conditions}
To examine tasks with different levels of imbalanced data, namely different target-background ratios, the participants were given two aerial image segments, each containing the same number of targets but differing in total area. For the \textit{dense-target task}, a segment with a total area of 0.15 km² and a high density of objects was selected. For the \textit{sparse-target task}, a segment with a total area of 4 km² was chosen (see Figure \ref{setup}B).
\subsubsection{Prior Experience}
\begin{figure}[h]
    \centering
    \includegraphics[width=1\linewidth]{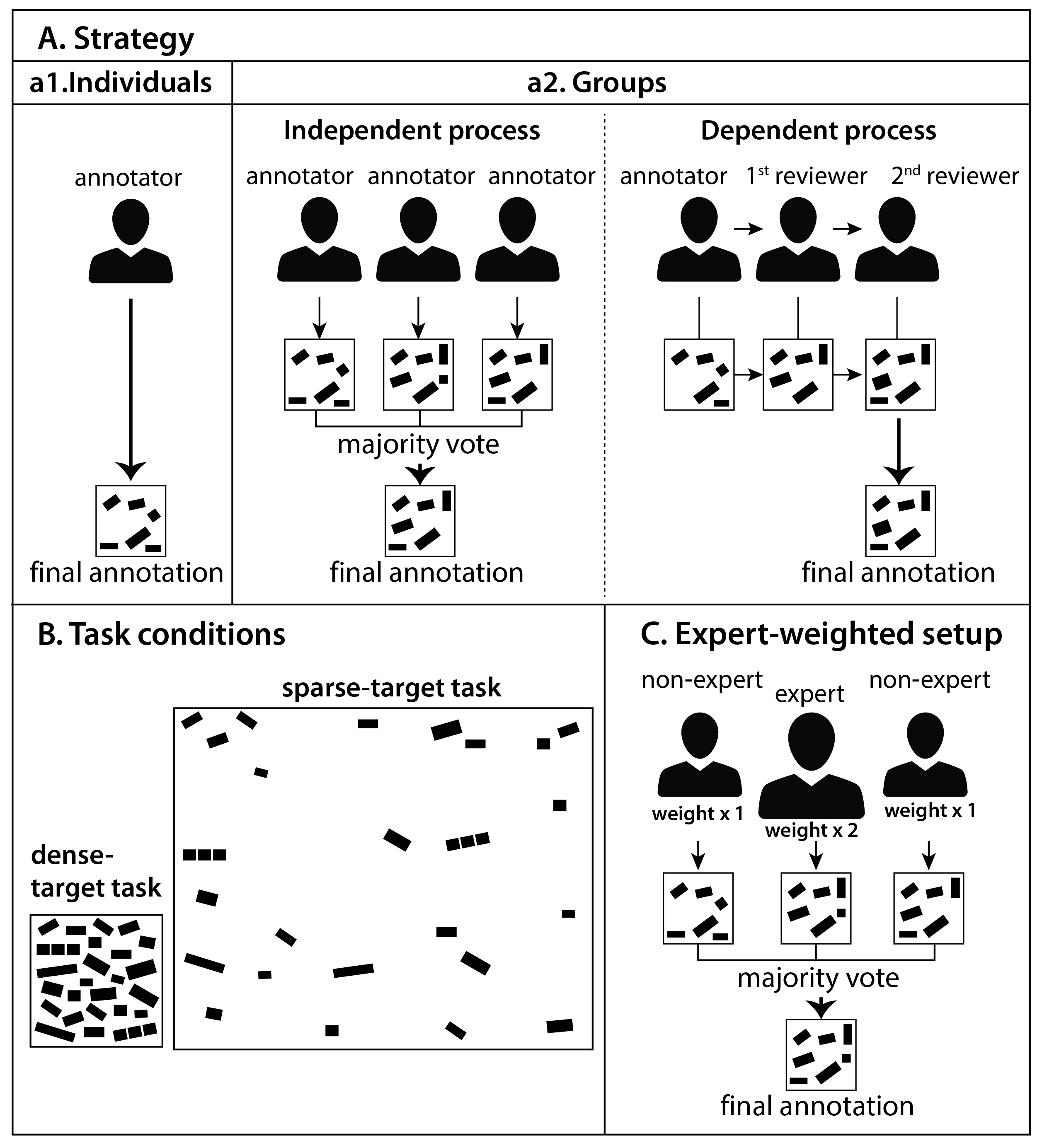}
    \caption{Experimental setup overview: A – Strategy setup: Individuals annotator (a1) versus groups of 3 annotators (a2). Within the groups: Independent Process (a2 on the left) - Each annotator creates an annotation separately. The final annotation is determined by majority vote. An object marked by at least 2 annotators will be included in the final annotation; Dependent Process (a2 on the right) - The first annotator passes the annotation to a reviewer who corrects it and passes the corrected product to a second reviewer, who finalizes the annotation. B – Dense-target task versus sparse-target task. Each task contains the same number of targets spread over different area sizes for varying target-background imbalance. C – Expert-weighted setup: assigning double weight to the expert annotator in the group compared to the non-expert annotators. This setup is compared to an unweighted setup (a2 left panel).}
    \label{setup}
\end{figure}

In addition to the performance comparison between individual expert and non-expert annotators, we designed a setup in which the expert annotator was given double the weight when voting on the objects to be included in the final annotation (see Figure \ref{setup}C). This setup was applied to teams conducting an independent annotation process.

\subsection{Experimental Analysis}
\subsubsection{Data}
RS image: An aerial photograph of the northern Negev of Israel (centered at 30\degree 40\textquotesingle{} N, 34\degree 50\textquotesingle{} E)  \cite{bluestein2023economic} from 2020, with a resolution of 0.15 meters, was selected for the experiment. In this area, over 1,300 clusters of Bedouin settlements are scattered across more than 600 square kilometers. This population is disconnected from the national electricity grid and is therefore characterized by high adoption rates of small household solar panels, located both on the ground and on rooftops, positioned at various angles and locations.

Golden standard dataset: The performance evaluation of the annotators was conducted by comparison to a gold standard. The gold standard was established through a two-step process. In the first step, the aerial image segments were annotated by an expert in the interpretation and annotation of RS images and reviewed by another annotator. In the second step, another independent review of the annotation was carried out by four RS experts who are deeply familiar with the study area. The gold standard was accepted only after achieving a complete consensus among the experts. 

\subsubsection{Performance Evaluation Metrics}
Across all setups and tasks, a comparison using a confusion matrix was made between the final annotations of each annotator/team and the gold standard. 

\begin{figure}[h!]
    \centering
    \includegraphics[width=1\linewidth]{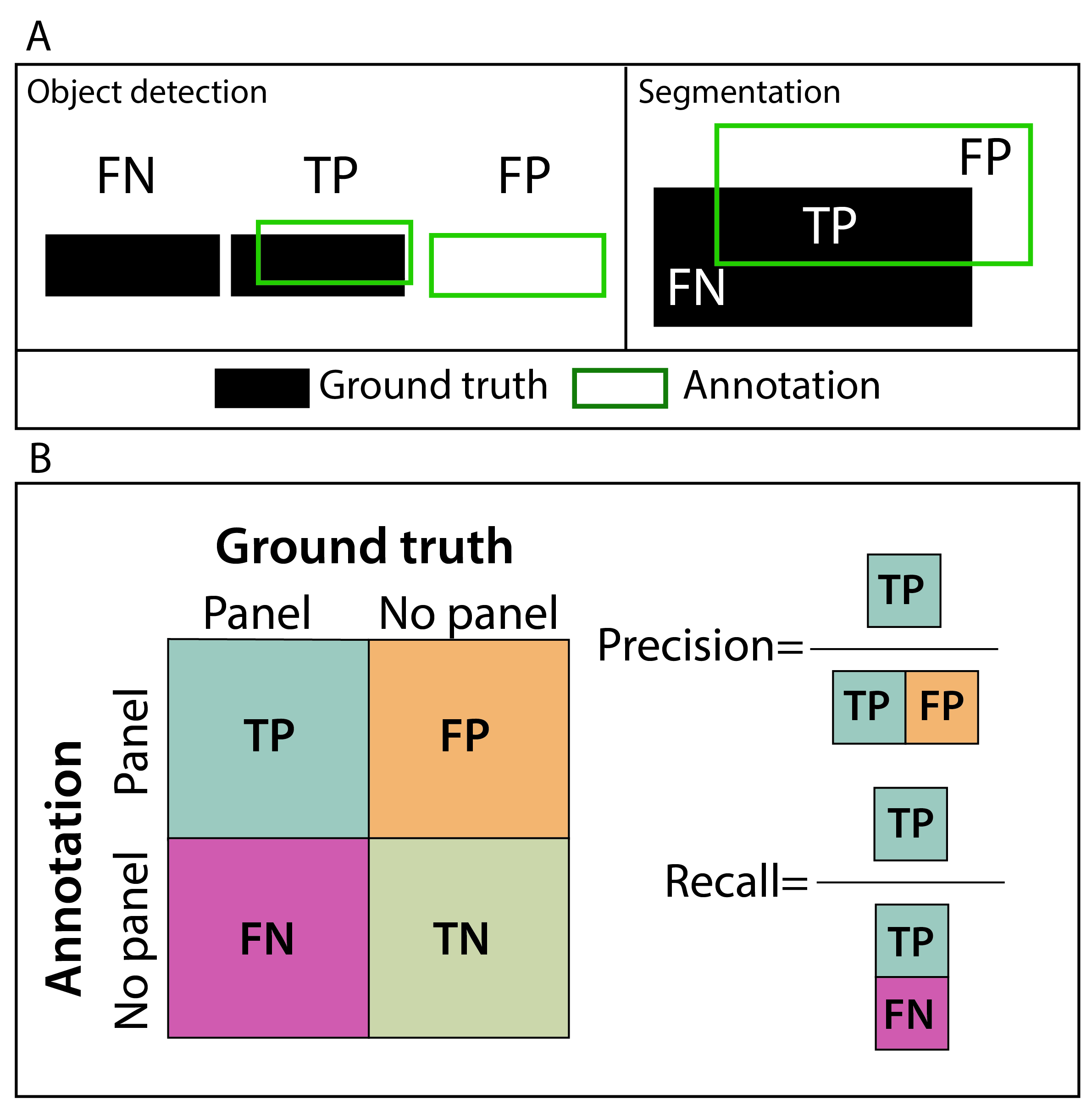}
    \caption{Confusion matrix and performance metrics in OD and segmentation: A - Components of confusion matrix for performance evaluations. On the left: Matrix components for OD. An object is defined as a True Positive if it has at least 60\%  overlap with ground truth. The matrix components quantify the number of annotated objects (panels) in each category. On the right: Matrix components for segmentation. The components represent the number of annotated pixels in each category. B – on the left: A confusion matrix with a tool to represent all combinations between ground truth and annotation in binary classification. On the right: Performance metrics derived from the confusion matrix. \textit{Precision} measures the ratio of correctly identified panels to all annotated panels, representing the accuracy of positive annotations; \textit{Recall} measures the ratio of correctly identified to all actual panels, representing the model's ability to identify all relevant instances.}
    \label{CM}
\end{figure}

\begin{figure*}[]
    \centering
    \includegraphics[width=1\textwidth]{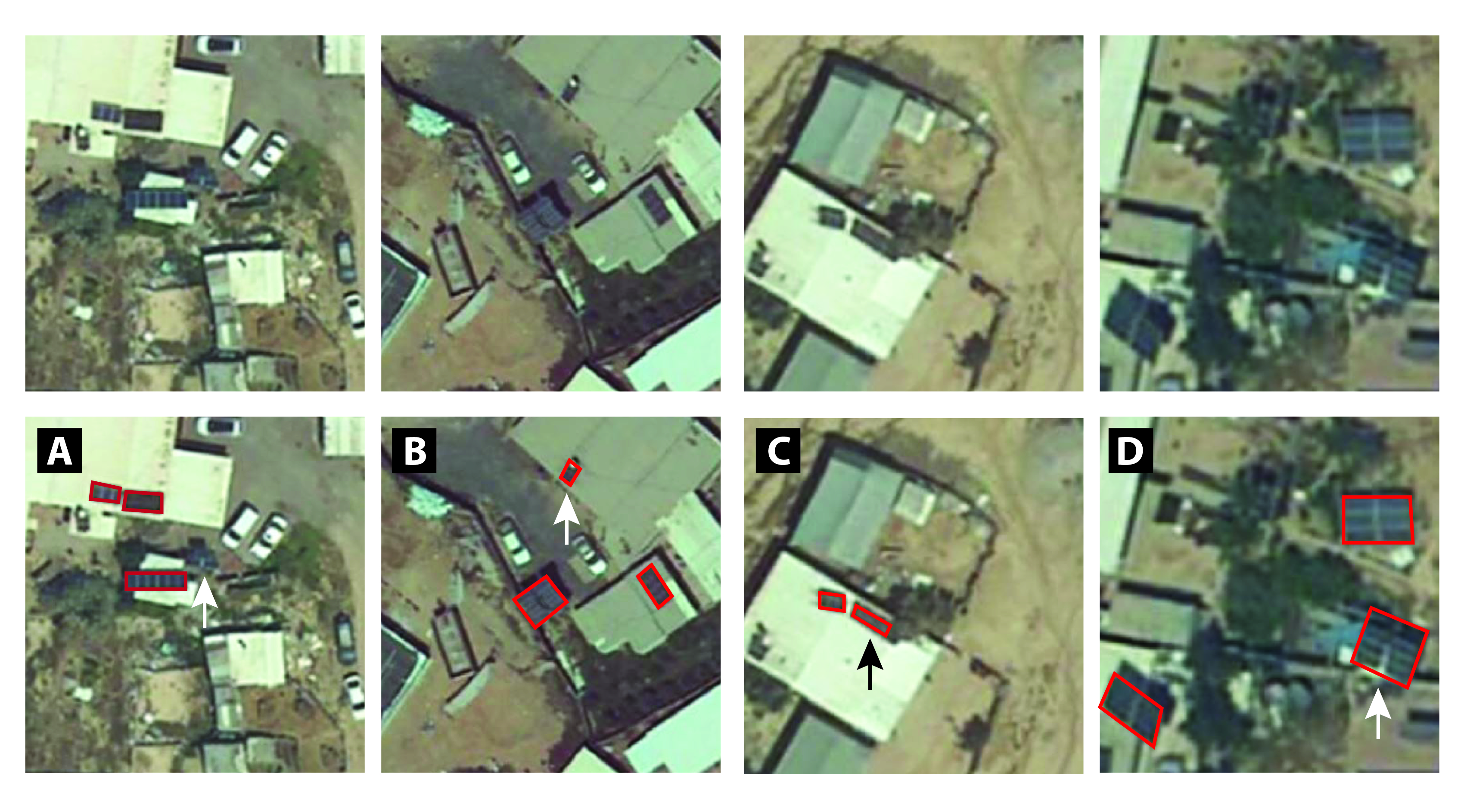}
    \captionsetup{width=\textwidth}
    \caption{Top row: Examples from the annotation task. Annotators were asked to identify and segment solar panels. Bottom row: Examples of annotations (red rectangle); A – an unannotated panel (FN object); B – wrong detection (FP object), where the annotated object is a sun-heated boiler; C – under-segmentation (FN pixels), with the panel not fully annotated; D – over-segmentation (FP pixels), where the annotation includes the shadow of the panel.}
    \label{examples}
\end{figure*}

\textit{A confusion matrix} is a table that provides a detailed summary of a labeling performance by cross-referencing annotated and actual labels, illustrating the counts of true positives (TP), true negatives (TN), false positives (FP), and false negatives (FN), thereby serving as the basis for calculating evaluation metrics, such as accuracy, precision, and recall (Figure \ref{CM}). Examples of the performed annotations and annotation errors are presented in Figure \ref{examples}. The detection and delineation of solar panels is an imbalanced data problem because the target-background ratio can be extremely low. Therefore, the \textit{accuracy} metric is not suitable for performance evaluation due to the accuracy paradox \cite{10.1371/journal.pone.0084217}. The performance evaluation focused on two metrics that are appropriate for imbalanced problems: \textit{Pecision} and \textit{recall}.

Precision quantifies the proportion of correctly identified target instances out of all instances identified as targets, reflecting the annotator's ability to minimize false positives and ensure the relevance of its annotations.

\begin{equation}
\begin{aligned}
\text{Precision} &= \frac{TP}{TP + FP} \\
\end{aligned}
\label{Precision}
\end{equation}
\setlength{\abovedisplayskip}{0pt}
\setlength{\belowdisplayskip}{3pt}

The complement of precision is the false positive rate (Type I errors), which reflects the likelihood that a positive annotation made by the annotator is actually incorrect.

Recall measures the proportion of actual positive instances correctly identified by the annotator, capturing its ability to detect all relevant cases (true positives) and minimizing false negatives.

\begin{equation}
\begin{aligned}
\text{Recall} &= \frac{TP}{TP + FN} \\
\end{aligned}
\label{Recall}
\end{equation}
\setlength{\abovedisplayskip}{0pt}
\setlength{\belowdisplayskip}{3pt}

The complement of recall is the false negative rate (Type II errors), which indicates the proportion of positive instances that were missed by the annotator, highlighting the likelihood of failing to identify a true positive.

The F1 score is the harmonic mean of precision and recall and provides a balanced view of performance, especially when there is a trade-off between precision and recall, making it particularly suitable for evaluating performance where one type of error is not more significant than the other.
\setlength{\abovedisplayskip}{0pt}
\setlength{\belowdisplayskip}{3pt}

\begin{equation}
\begin{aligned}
\text{F1 Score} &= 2 \times \frac{\text{Precision} \times \text{Recall}}{\text{Precision} + \text{Recall}} \\
\end{aligned}
\label{F1 Score}
\end{equation}

\subsubsection{Spatial Data Processing}
When combining the polygonal layers of participants of a given group (Figure \ref{setup}) into their final prediction, the intersections of at least two participants were considered. In the expert-based setup, expert participants were given double votes, i.e., their polygonal layer was duplicated (Figure \ref{setup}C). Predicted "panels" with areas less than 1 $m^2$ were discarded. Then, agreement metrics between the combined participant layer and the reference ("gold standard") layer were calculated. In the OD approach (Figure \ref{CM}A), we considered polygon count, where group polygons covering more than 60\% of a reference polygon they intersect with are considered TP, while the remaining "unmatched" group and reference polygons are considered FP and FN, respectively. In the segmentation approach, we considered the area (in $m^2$) of overlap (Figure \ref{CM}B), rather than polygon count. The process was repeated for the different participant groupings, tasks, and expert treatments (Figure \ref{setup}), while the reference layer remained fixed. Spatial data processing was done in R version 4.4.1 \cite{r_core_team} and package \textit{sf}\cite{r_sf}. 

\subsubsection{Statistical Analysis}
Performance comparison across all setups and tasks was conducted using a T-test for differences in means of independent samples. A preliminary test for homoscedasticity was performed using the F-ratio to determine the method for variance estimation. For a visual representation of the results of each comparison, we used a box and whisker plot that summarizes the distribution of a dataset and its central tendency.

\section{Results}
To address the four operative objectives (OBJ) outlined at the end of the Introduction, we formulated the following eight research questions (RQ). Each RQ is assigned with one of the four BJs.

\textit{OBJ1:} To compare the performance of human annotators in OD versus segmentation tasks.
\begin{list}{}{\leftmargin=1em}
\item \textbf{RQ1:} Do annotators achieve higher recall, precision, and F1 scores in OD compared with segmentation?
\end{list}
\textit{OBJ2:} To analyze distinctions between Type I errors (False Positives) and Type II errors (False Negatives) in annotations \textit{across} annotation setup strategies and task conditions.
\begin{list}{}{\leftmargin=1em}
\item\textbf{RQ2:} Is there a difference between precision (indicating \textit{correct} detection) and recall (indicating \textit{complete} detection) in both OD and segmentation tasks when comparing individual with group-based annotation setup?

\textbf{RQ3:} Is there a difference between precision and recall in both OD and segmentation tasks when comparing independent group annotation (based on majority voting) with dependent group annotation (involving sequential reviewing)?

\textbf{RQ4:} Is there a difference between precision and recall in tasks where target-background ratio is higher (with objects clumped in small areas) compared with tasks having lower target-background ratios (with objects dispersed over wide areas)?
\end{list}

\textit{OBJ3}: To compare annotators' performance across annotation setup strategies and varying task conditions.  
\begin{list}{}{\leftmargin=1em}
\item \textbf{RQ5:} Is there a difference, both in precision and recall, for OD and segmentation tasks, when comparing an independent to a dependent setup?

\textbf{RQ6:} Is there a difference, both in precision and recall in OD and segmentation tasks, between a task with a low target-background ratio and a task with a high target-background ratio?
\end{list}

\textit{OBJ4}: To examine the impact of prior experience in RS data interpretation, digitization, and annotation on annotation performance.

\begin{list}{}{\leftmargin=1em}
\item\textbf{RQ7:} Is there a difference in OD and segmentation precision, recall, and F1 between expert and non-expert annotators in an individual annotation setup?

\textbf{RQ8:} Do the precision, recall, and F1 scores in OD and segmentation tasks improve when expert annotations are given double-weighting in an independent annotation process?
\end{list}

\begin{figure}[h!]
    \centering
    \includegraphics[width=1\linewidth]{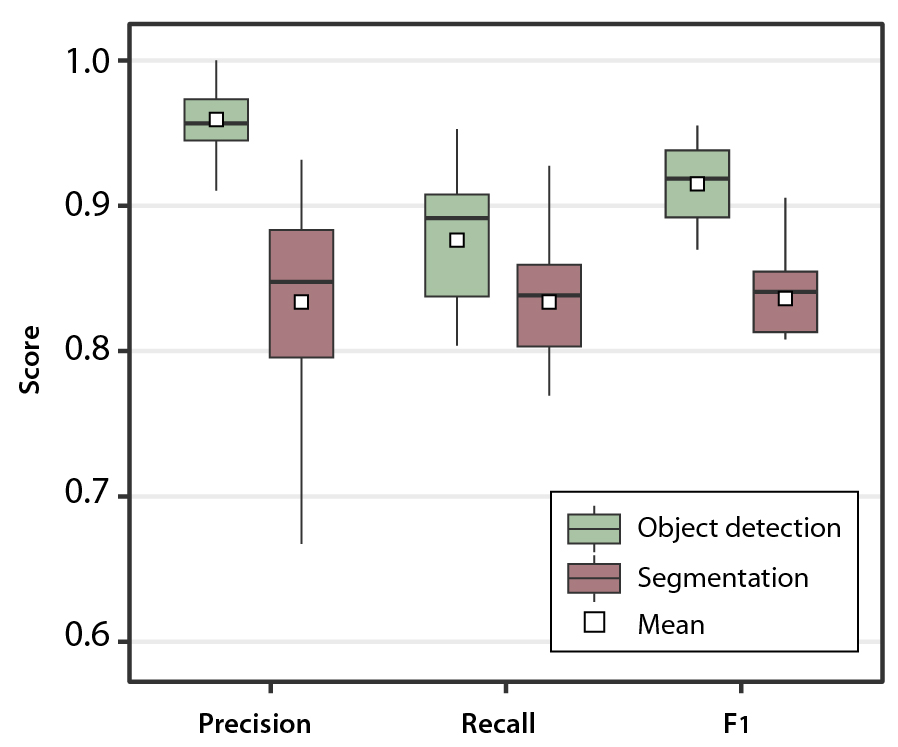}
    \caption{Comparison of evaluation metrics between OD and segmentation tasks. All evaluation metrics are higher in OD compared with segmentation, indicating superior performance in identification compared with accurate delineation. Note that differences between the tasks are more pronounced in precision.}
    \label{od_seg}
\end{figure}

\subsection{Object Detection versus Segmentation}
To address RQ1, the evaluation metrics: precision, recall, and F1 score, were compared between OD and segmentation tasks, without distinguishing between task conditions or experimental setups. As shown in Figure \ref{od_seg}, performance in OD is higher across all three metrics compared with segmentation. However, differences in precision are more pronounced (0.13, p-value for mean differences \textless{0.000}) compared with recall (0.04, p-value for mean differences \textless{0.009}). The difference in overall average performance (F1) between the tasks is 0.08 (p-value for mean differences \textless{0.000}).

\subsection{Differences in Error Types}
\begin{figure}[htbp!]
    \centering
    \includegraphics[width=1\linewidth]{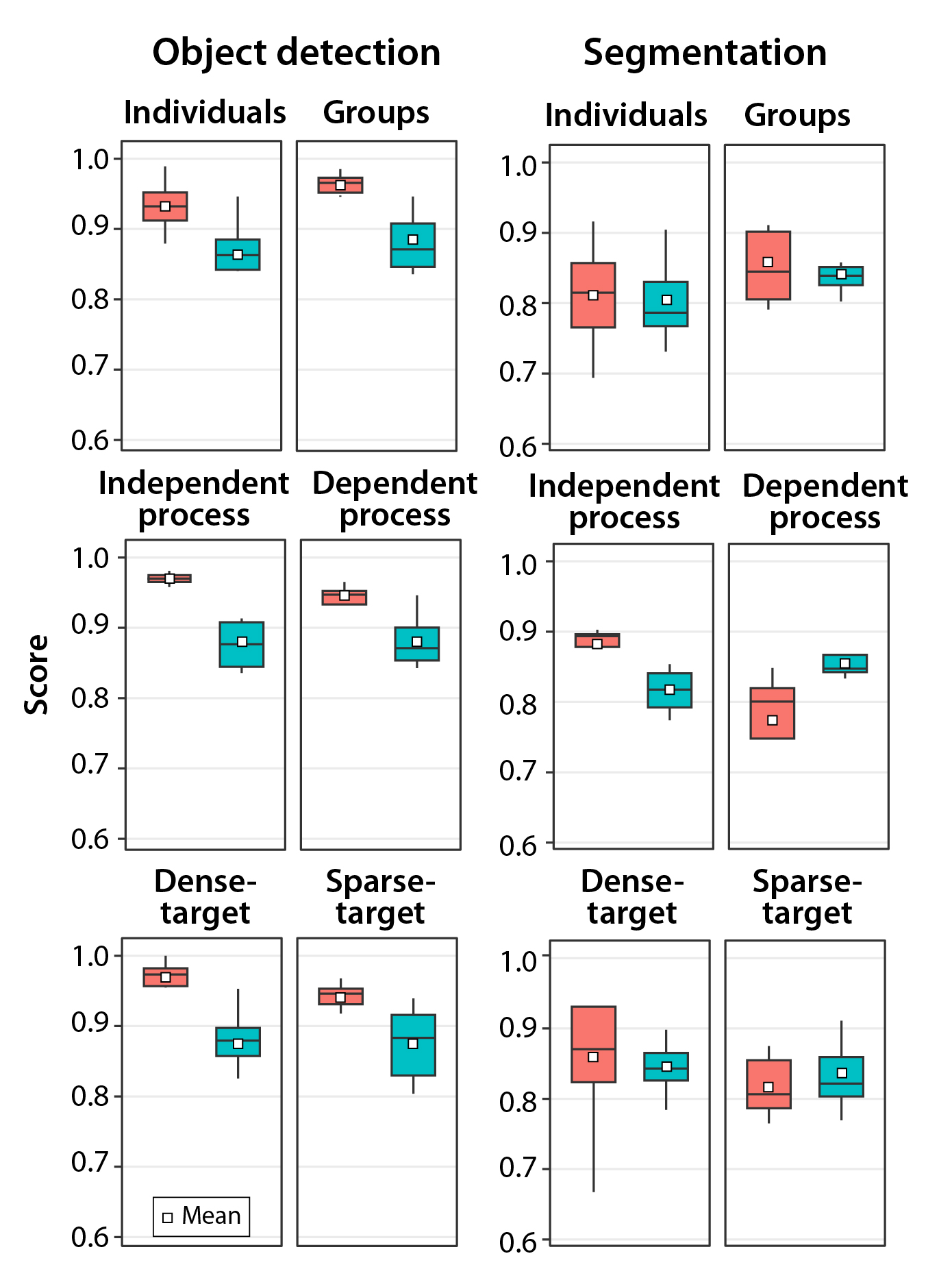}
    \caption{Comparison of Precision and Recall Results. In OD, across all setups and tasks, precision is higher than recall. In segmentation, the results are mixed: in the weighted-expert setup and the dependent annotation process, recall is higher than precision. In the independent annotation process, precision results are higher than recall results. The t-test results for mean differences are highlighted in Table \ref{ttest_pr}.}
    \label{p-r}
\end{figure}

To achieve the second objective, differences between precision and Recall were analyzed. As shown in Figure \ref{p-r}, in the OD task, precision scores are higher than recall scores across both individual and group annotation setups (RQ2) (differences of 0.05 and 0.07, respectively), independent and dependent annotation processes (RQ3) differences of 0.06 and 0.10, respectively), and dense-target as well as in sparse-target tasks (RQ4) (differences of 0.06 and 0.09, respectively). In all scenarios, the differences in average evaluation metrics are statistically significant (see Table \ref{ttest_pr}) indicating higher rates of Type II compared with Type I errors. In the segmentation task, no significant differences were found between the two evaluation metrics in any of the comparisons, except in the independent and dependent setups, where in the latter, recall results surpass precision (difference of 0.09), indicating a higher rate of Type I errors within the dependent setup and a tendency to overestimate the delineation. 

\begin{table}[h]
\renewcommand{\arraystretch}{1.2}
\begin{tabular}{lll m{1.3cm}} \hline
\multicolumn{2}{l}{}                                        & \textbf{OD} & \textbf{Segment.} \\\hline
\multirow{2}{*}{\textbf{Task conditions}} & Dense                 & 0.0001***                   & 0.4548                \\
                                    & Sparse                & 0.0038**                    & 0.1997                \\\hline
\multirow{4}{*}{\textbf{Setup strategy}}  & Individuals           & 0.004*                      & 0.0765                \\
                                    & Groups                & 0.001**                     & 0.2907                \\
                                    & Independent           & 0.0018**                    & 0.0144**              \\
                                    & Dependent             & 0.0434*                     & 0.036*                \\\hline
\textbf{Expertise effect}         & Expert-weighted setup & 0.0111*                     & 0.0241*   \\\hline            
\end{tabular}
\caption{P-values of t-tests for differences in mean precision and recall across setup strategies, task conditions, and expert-weighted setup. Significant differences between precision and recall were observed in all OD setups and tasks. In the case of segmentation, significant distinctions between precision and recall were found both in the independent and dependent annotation processes, and in the expert-weighted setup.}
\label{ttest_pr}
\end{table}

  \subsection{Performance Comparisons}
\begin{figure*}[h]
    \centering
    \includegraphics[width=1\textwidth]{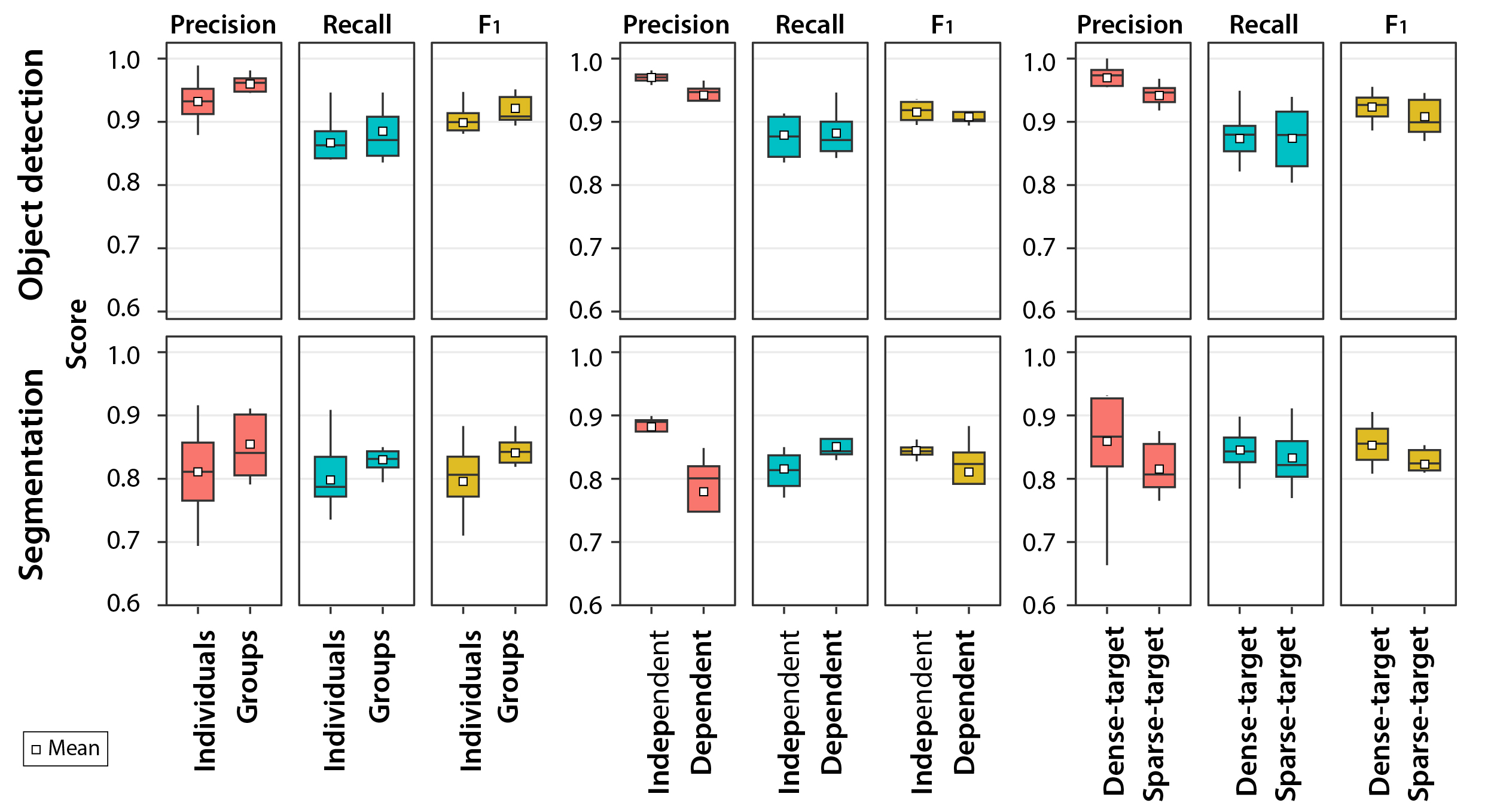}
    \captionsetup{width=\textwidth}
    \caption{Performance metrics comparison across task setups and conditions: top row – OD; bottom row – segmentation. In OD, precision is higher in the task of clumped objects compared with the sparse objects; among groups compared with single annotators; and in the independent annotation process compared with the dependent process. No differences were found between unweighted and weighted expert setups. Additionally, no significant differences in recall were observed across all setups and tasks (see Table \ref{ttest_performance} for t-test results). In the case of segmentation, significant differences were found across all task setups and conditions, with higher precision in task of clumped objects, among groups, and in the case of independent annotation process. Note that the weighted expert setup achieved lower precision compared with unweighted setup. Differences in recall were found only between single annotators and groups, with groups achieving higher scores.}
    \label{performance}
\end{figure*}

To achieve the third objective, a comparison was made between the evaluation metrics across various strategy setups (RQ5) and task conditions (RQ6), both in OD and segmentation tasks. As indicated in Figure \ref{performance}, higher precision was achieved in both OD and segmentation in group setups compared with individual annotators setups (difference of 0.035 and 0.015, respectively). Similar results achieved in the independent annotation process compared with the dependent process (difference of 0.03 and 0.12, respectively), and in tasks with a high target-background ratio (dense-target) compared with tasks with a low target-background ratio (sparse-target) (difference of 0.04 and 0.06, respectively), with a lower rate of Type I errors in these scenarios. In all these cases, the superiority of precision over recall was statistically significant (see Table \ref{ttest_performance}). Note the large difference in precision scores between the independent/dependent setups in the segmentation task, which stems both from a below average rate of Type I errors in the independent setup and an above average rate of Type I errors in the dependent setup.  

In the OD task, no significant differences in recall were found across any of these scenarios, with the rate of Type II errors remaining consistent across all setups and task conditions. In segmentation, however, significant differences in recall were found between the dependent and independent setups, where the dependent setup reduced the rate of Type II errors compared with the independent setup and decreased the extent of under-segmentation in target delineations.
  
\begin{table*}[h]
\captionsetup{width=\textwidth}
\renewcommand{\arraystretch}{1.2}
\begin{tabular}{m{2cm}m{6cm}|m{1.1cm}m{1.1cm}m{1.1cm}|m{1.1cm}m{1.1cm}m{1.1cm}}
\hline
\multicolumn{2}{l|}{\multirow{2}{*}{\textbf{}}}                                             & \multicolumn{3}{c|}{\textbf{Object detection}} & \multicolumn{3}{c}{\textbf{Segmentation}} \\\cline{3-8}
\multicolumn{2}{l|}{}                                                                       & Precision       & Recall       & F1           & Precision      & Recall      & F1         \\\hline
\textbf{Task conditions}                         & Dense   versus Sparse                         & 0.0434*         & 0.3847       & 0.1223       & 0.0323*        & 0.2359      & 0.106      \\\hline
\multirow{2}{*}{\textbf{Setup strategy}}         & Individuals   versus Groups                   & 0.001**         & 0.391        & 0.042*       & 0.02*          & 0.041*      & 0.01**     \\
                                           & Independent versus Dependent                  & 0.0476*         & 0.4109       & 0.262        & 0.0312*        & 0.0819      & 0.1485     \\\hline
\multirow{2}{*}{\textbf{Expertise effect}} & Experts   versus Non-expert (Individuals)     &   0.442              & 0.208             &   0.323           &    0.47            &  0.48           &  0.451          \\
                                           & Weighted-expert setup versus Unweighted setup & 0.1035          & 0.145        & 0.2275       & 0.0069**       & 0.0711      & 0.0288 \\\hline   
\end{tabular}
 \caption{P-values from t-tests showing differences in performance metrics across strategy setups, task conditions, and expertise. Significant differences in precision are observed between task conditions and strategy setups in both OD and segmentation — distinguishing between sparse- versus dense-target tasks, between individual annotators versus groups, and between independent versus dependent annotation processes. Significant differences in recall were also noted in segmentation between individuals and groups.}
\label{ttest_performance}

\end{table*}

\subsection {Impact of Prior Experience}
The impact of prior experience in interpretation, digitization, and annotation of RS tasks on annotation performance (OBJ4) was examined in two ways: \textit{First}, a comparison was made between the evaluation metrics of individual expert annotators (annotators with 22 months experience in average) versus those of annotators without prior experience (non-experts) (RQ7). \textit{Second}, we investigated whether the results of group annotations were improved when the expert's annotations were given a double weight in the majority vote decision compared with non-expert group members (RQ8). 

\begin{figure}[h!]
    \centering
    \includegraphics[width=1\linewidth]{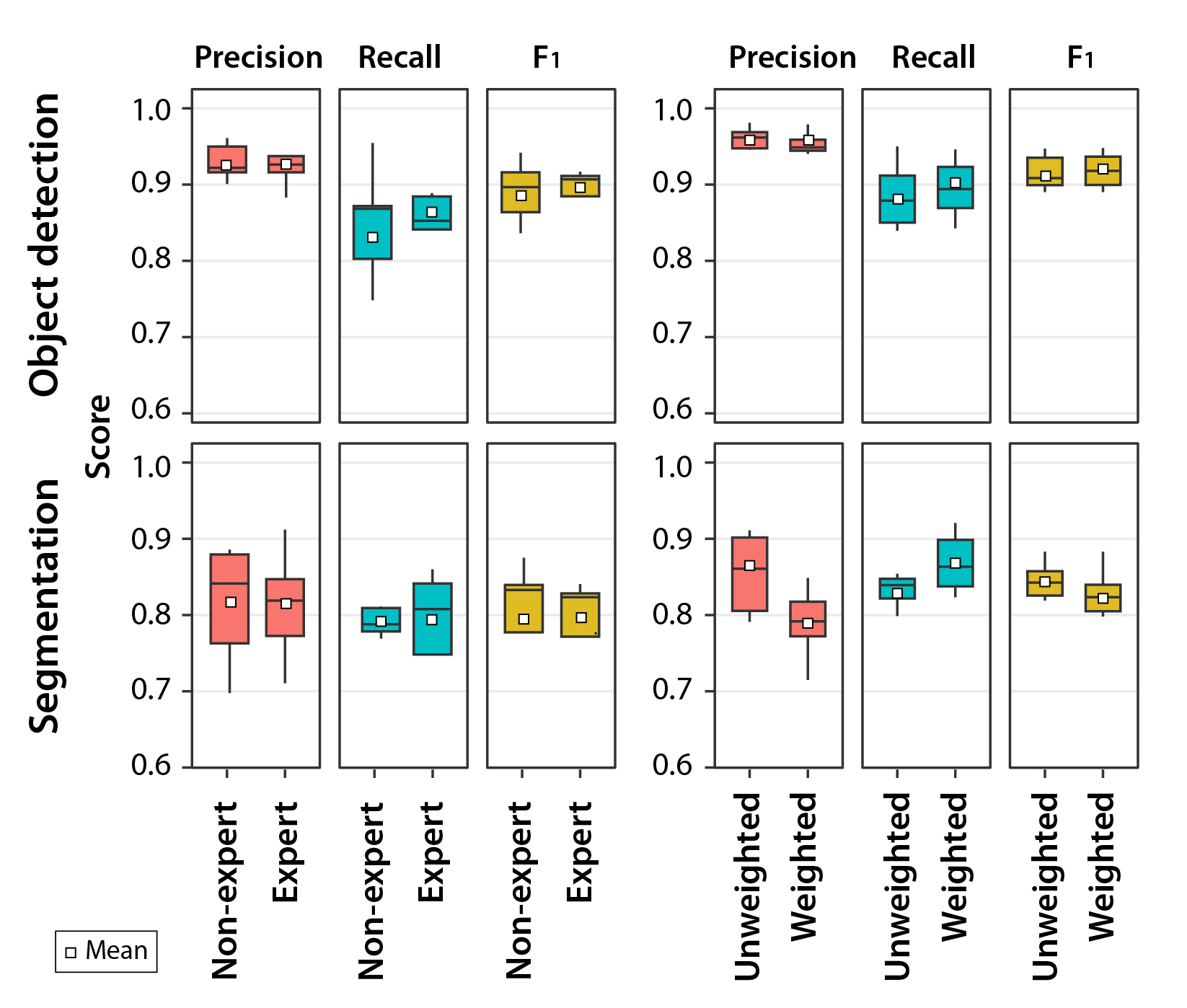}
    \caption{Performance comparison for impact of expertise assessment: On the left - evaluation metrics of individual non-experts versus experts. No significant difference was found in annotation performance in either OD or segmentation tasks. On the right - evaluation metrics for the unweighted setup versus the weighted setup. A significant difference was found in precision, with the weighted setup showing lower performance compared to the unweighted setup.}
    \label{performance_exp}
\end{figure}

A comparison of the performance of expert/non-expert individual annotators reveals that there are no significant differences between the two groups across all evaluation metrics, both in OD and segmentation tasks (see Table \ref{ttest_performance} and Figure \ref{performance_exp}). As observed in other scenarios, precision is significantly higher than recall for both experts and non-experts (P-values for the t-test of mean differences are 0.0235 and 0.0162, respectively).

\begin{figure}[h!]
    \centering
    \includegraphics[width=1\linewidth]{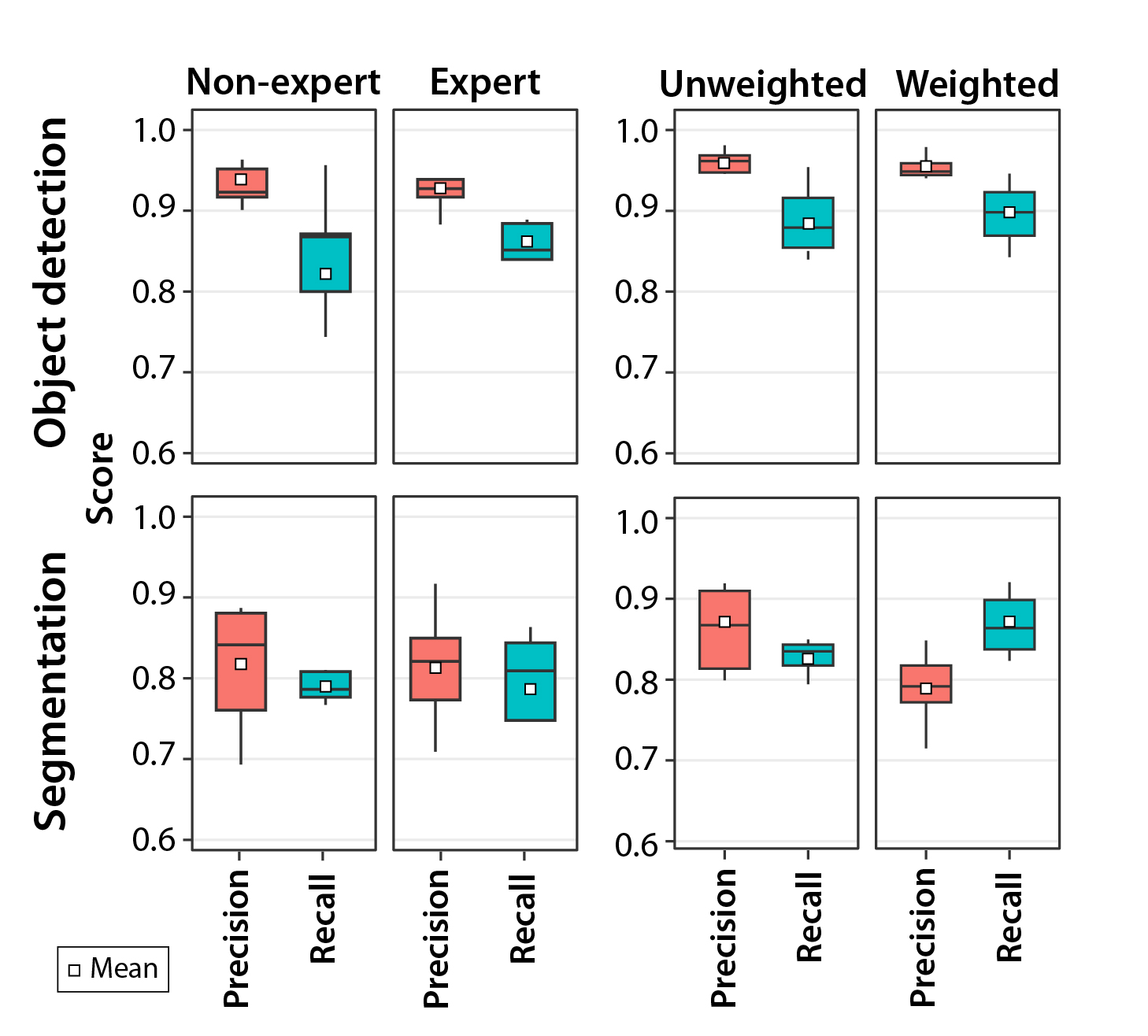}
    \caption{Comparison of Precision and Recall Results between Experts and Non-Experts. Left panel: Results for individual non-experts and experts. Right panel: Expert-weighted setup vs. unweighted setup. In OD, there is a clear difference between the evaluation metrics, with precision being notably higher than recall. In segmentation, no significant differences were observed, except in the weighted setup, where precision is lower than recall, indicating a tendency toward over-estimation in target delineation.}
    \label{p_r_exp}
\end{figure}

As shown in Figure \ref{performance_exp}, in the OD task, a double weighting assigned with expert annotators did not impact annotation performance and did not reduce error rates (no significant difference in the evaluation metrics—see Table \ref{ttest_performance}). In the segmentation task, there is a significant difference in precision (0.10 in favor of the unweighted setup), with the weighted setup leading to lower performance, increasing Type I errors, indicating a tendency toward overestimation in target delineation.
In addition, notice that also in the weighted setup, a significant difference was observed between precision and recall (see Figure \ref{p_r_exp} and Table \ref{ttest_pr}). However, in the OD task, precision scores are higher than recall (difference of 0.05), consistent with the other setups and conditions tested, whereas, in the segmentation task, precision scores are lower than recall (difference of 0.08) and point to the experts' propensity for target-size overestimation.

\section{Discussion}

\subsection{Higher Performance in OD Compared with Segmentation}
A significant difference was observed in the annotator's performance with greater success in OD than in segmentation, both in terms of accuracy (precision) and completeness (recall). Lower performance in segmentation was reported also by studies examining annotation quality in standard ground photography. For example, \cite{Benenson_2019_CVPR} showed an average accuracy of 75\% in the boundary accuracy of various objects from the COCO dataset (one of the largest image corpus for computer vision tasks training) \cite{lin2014microsoft}. The higher average performance of 83\% we observed could be attributed to the skill of the participants, who are geoinformatics students and may have an affinity towards RS tasks more than the general population has to a random ground photo.
The difference in performance in OD versus segmentation can be explained by the following: segmentation is a more complex task that requires higher cognitive demands, which inevitably results in lower performance. For example \cite{marr2010vision} found that in visual processing, OD involves a general pattern matching by the brain, while delineation requires detailed and precise boundary trace, which may increase cognitive load. Note that segmentation errors indicate the overall pixels that were over- and underestimated, and are not necessarily related to the number of detected objects. Note that the dataset panel size histogram is normally distributed.
Another independent reason can be the fact that segmentation requires different skills than OD, which are rooted in a different type of intelligence than demanded for OD. As \cite{gardner2011frames} distinguishes, in his seminal work on types and sub-types of intelligence, Spatial Intelligence-primarily required for OD-is closely related but a distinct concept from Visual-Perceptual Intelligence required for accurately delineating detailed polygons. Given the representative sample in our experiment (Kolmogorov-Smirnov Test, p-value\textgreater{0.05}), it is possible that prevalence of individuals with the first type of intelligence is higher than those with the second type. 
To the best of our knowledge, there are no existing RS studies that compare the performance between the two key computer vision tasks — OD and segmentation — even though human annotators in RS tasks are usually required to perform both simultaneously. The importance of this finding lies in understanding the need to coordinate between the annotator's personal skills and the task requirements. By aligning the right task's requirements, we can achieve faster, more accurate, and higher-quality results. It could also be useful to conduct an initial evaluation of the potential annotators for RS tasks using validated tests that assess the level of spatial intelligence, such as \textit{Raven's Progressive Matrices} \cite{raven1998raven} or \textit{Mental Rotation Test} \cite{vandenberg1978mental} for batter matching between the annotator's competencies and the task. 

Alongside, the emergence of large segmentation models, such as \textit{Segment Anything Model} by Meta \cite{10378323}, which allow defining user prompts such as raw masks, creates the possibility for using lower-quality annotation by humans that can be improved by machine segmentation \cite{rafaeli2024prompt}.

\subsection{Consistent Bias in Error Types}
We found a pronounced tendency by the participants to commit more Type II errors (False Negatives) than Type I errors (False Positives). This finding, reported for the first time for RS tasks, observed across all experimental setups and task conditions, as reflected in the significant difference between precision and recall, clearly favoring the former (see Table \ref{ttest_pr}). This indicates that human annotators perform better in correctly identifying objects or delineating their boundaries (with very few active errors in identifying a non-object as an object), but they are more prone to miss objects of interest. This bias aligns with the \textit{Prospect theory}, formulated by Nobel laureate Kahneman and his co-author Tversky \cite{eec14168-5714-3ca8-b073-d038266f2734}. The prospect theory suggests, among other assertions, that humans are biased to prevent losses more than they strive to achieve gains. This bias has been coined by Kahneman and Tversky as the \textit{Loss Aversion bias} and was validated by numerous empirical studies in the behavioral sciences since the 1980s (for example \cite{Abdellaoui_2007} \cite{dd54c72a-3949-3a49-956c-54726a8c3880}). More detailed, the cognitive tendency in decision-making under uncertainty, where aversion to losses outweighs the appeal of equivalent gains, may lead individuals to prioritize avoiding losses over potential rewards, as losses are perceived as more significant and distressing than gains are satisfying. The value function v(x), that dictates the aversion bias, as described by \cite{eec14168-5714-3ca8-b073-d038266f2734}, is defined as follows:

\[
v(x) = 
\begin{cases}
x^\alpha & \text{if } x \geq 0, \\
-\lambda (-x)^\alpha & \text{if } x < 0,
\end{cases}
\]

where:

\begin{itemize}
    \item \(x\) represents the change in value (gain or loss).
    \item \(\alpha\) is a parameter typically in the range \(0 < \alpha \leq 1\), reflecting sensitivity to gains and losses (commonly \(\alpha \approx 0.88\)).
    \item \(\lambda\) is the loss aversion coefficient 
\end{itemize}

The upper part of the expression refers to gains evaluation, and the bottom shows the evaluation of losses. The loss aversion coefficient (\(\lambda\)) captures the degree of loss aversion, with \(\lambda > 1\) indicating that losses "hurt" more than equivalent gains feel good. Empirical studies often find \(\lambda\) values between 2 and 2.5 (see, for example, \cite{Tversky_1992}).

\begin{figure}[htbp!]
    \centering
     \fbox{\includegraphics[width=0.85\linewidth]{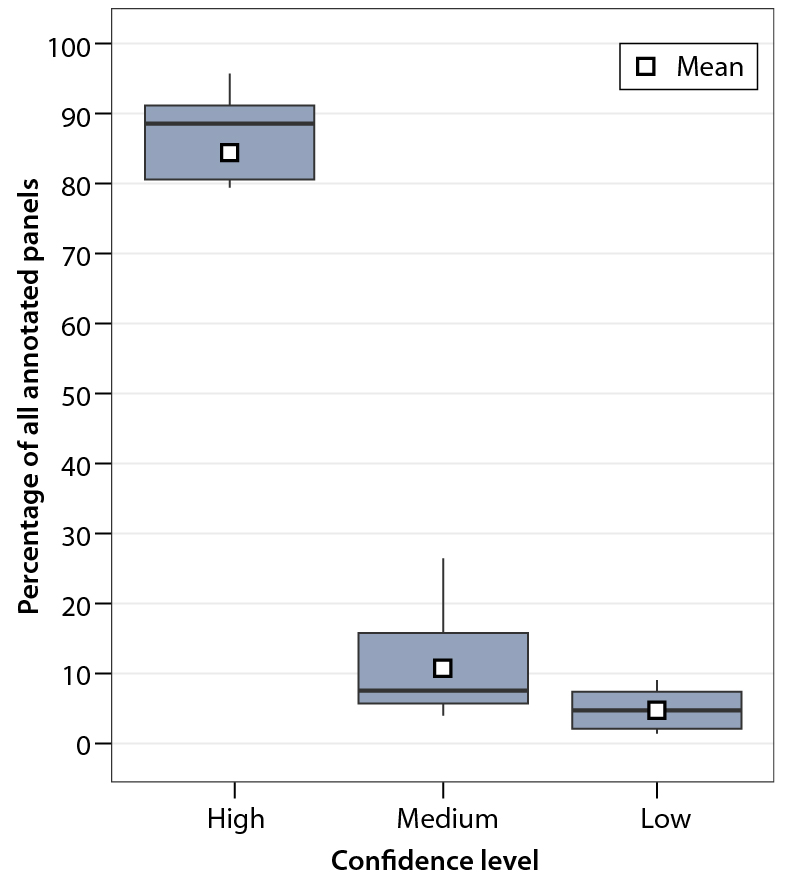}}
    \caption{Percentage of detected objects by reported confidence level: annotators were asked to rate the detected objects into three confidence levels indicating their certainty in object detected as a solar panel. Note that the vast majority of detected objects (84\%) were rated with high confidence level, while only 6\% on average were rated with a low confidence level.}
    \label{confidence}
\end{figure}

In the case of using data created by human annotators in RS, the loss aversion bias suggests that annotators are likely to refrain from marking an object they are uncertain about, thus risking a loss (missing a true object), rather than marking it and gaining an additional correct identification. This assertion aligns with our data on confidence levels in the correctness of identification reported by the participants during the experiment (Figure \ref{confidence}). The number of objects marked with high confidence is five-fold higher than those marked with medium or low confidence and accounts for 84\% of all labeled objects. This observation indicates a pronounced tendency to favor objects that the annotators are confident that they are indeed solar panels. 

In the segmentation task, significant differences in errors were found only in three cases, all in group setups: the dependent process; the independent process; and the expert-weighted. The reason for that may be because the cognitive mechanisms required for segmentation are fundamentally different than in OD, as was explained earlier in the Discussion Section. In OD, the participant considers the object as a generalized entity and inquires whether it is a panel or not. In the level of an object, the loss aversion bias is pertinent. In segmentation task, however, other skills of the participant are involved, such as exactness, ability to discern color variations, and comprehensive visual perception of the object. Therefore, an underestimation in delineation, that yields an exclusion of pixels, may not be caused by lost aversion. This may also be the reason why, in the dependent setup, the difference in errors flips: there is more overestimation than underestimation. This flip may reflect a tendency not to discern the boundaries of an object and to include shadows or parts of adjacent similar objects.

We suggest balancing this inherent bias observed here by designing an annotation setup that provides positive incentives for taking risks and marking objects with lower confidence levels in their validity. Another option is to guide the annotators to mark objects with medium or even low certainty, and these objects can be reevaluated in a subsequent stage by others. Such frameworks could reduce underestimation and lead to overall improved performance. 

\subsection{Annotator's Performance Across Strategies and Conditions} 

\subsubsection{Majority Vote Outperforms Reviewing Process}
We compared two annotation setups (see Figure \ref{setup}: (1) individual and group-based annotations; and (2) an independent group setup, where the final annotation is determined by a majority vote of independent annotators, and a dependent group setup, where a single annotator is followed by a sequential double control procedure by other annotators.

We found that the group setup outperformed the individual setup (reduction of 4\% and 1\% in Type I errors in the group setup in OD task and segmentation task, respectively), and the independent setup achieved significantly higher results than the dependent setup, most remarkably in the segmentation task (reduction of 3\% and 12\% in Type I errors in the group setup in OD task and segmentation task, respectively).
While it is not surprising that teams outperform individuals, regarding the group setup, our findings align with Occam's Razor, suggesting that the simple, straightforward majority vote strategy, harnessing the wisdom of the crowd, is more effective in achieving accurate annotations than the process carried out under double control procedure.

The increase in Type I errors in the dependent setup can be explained by a tendency towards overestimation by reviewers. The reviewer's task differs from that of the annotator because they receive a rather high-quality product (F1 score of 0.89 in OD task and 0.79 in segmentation for individual annotators) that they need to refine. However, the reviewer may be biased by the previous annotation, influencing their decision-making. For example, in an effort to justify their role, the reviewer may tend to mark objects they are less confident about, which could potentially lead to increased overestimation in the final annotations. The reviewer may also hesitate to remove labels made by a previous annotator.

Based on these findings, we recommend prioritizing a team setup of independent annotators who perform the same task in parallel. Increasing the number of annotators in the team is likely to enhance the reliability of the final outcome.

It is important to note that no difference was observed in Type II errors between the group setups, namely, the independent process did not succeed in reducing the rate of Type II errors compared with the dependent process. This is likely because even a group of annotators fails to identify obscured objects that are difficult to detect due to low contrast, shading, resemblance, or close adjacency to other objects, etc.

\subsubsection{The effect of imbalanced data}
We examined the impact of the target-background ratio on the performance of the annotators. In both OD and segmentation, The results indicate significantly higher performance in more balanced tasks, with target spread over a smaller area (higher target-background ratio).

We suggest the following explanations for the observed differences in performance. First, annotating large regions with sparsely distributed targets may reduce performance by increasing the search time across vast, often monotonous areas, which can be tiring for annotators, and lead to missed detection (false negative). Conversely, searching through extensive background regions can increase the statistical likelihood of false positives. Our results support the latter explanation, as they showed that in the imbalanced conditions, the rate of Type I errors was higher by 3.5\% compared with the more balanced task. Lastly, in more balanced tasks, where multiple instances of the same entities are visible within a small and condensed space, annotators can become more familiar with the objects and make comparisons within the same field of view, thereby improving their ability to use the surrounding context and correctly identify targets.

Based on our findings - for tasks with a low target-background ratio, we recommend dividing the space into smaller, more manageable tasks, which can enhance annotation performance by increasing the target-background ratio.

\subsection{No Advantage in Performance for Experts} 
We examined whether there are performance differences between annotators without specific annotation training and those with an average of 22 experience months in RS image interpretation and annotation. In both setups — comparing individual experts to non-experts and giving experts double weight in the majority vote strategy — no significant differences were found between the annotator's performances, and additionally they both have more type II errors than type I. 

Moreover, in the segmentation task, assigning double weight to experts even increased the overall overestimation. This observation can be explained by overconfidence among experts, which refers to their excessive confidence in their ability to detect subtle features. Consequently, they may identify and segment, in our case, more areas as part of the object than necessary, ultimately resulting in over-segmentation. This tendency of experts toward overconfidence has been validated in numerous behavioral sciences studies investigating decision-making in uncertainty, including those of \cite{tversky1974judgment}, \cite{sniezek2001trust} and \cite{griffin1992weighing}, which explore cognitive biases among experts. Lower performances among experts could be also attributed to their tendency to perform these tasks more automatically due to their broad span of practice. This automation can lead to a devotion of less attention or conscious thought to the task, potentially impacting the accuracy of the annotation.
We found no advantage in preferring expert annotators over non-expert annotators in RS tasks of object detection and segmentation. Comprehensive training and familiarity with aerial imagery and basic digitization tools are sufficient to yield high performance in these tasks.

\section{Conclusion}
This study presents an experiment evaluating the performance of human annotators in RS segmentation and OD tasks, with a particular focus on examining differences in error types across various strategy setups and task conditions. The results indicate that human annotators generally perform better in OD than in segmentation tasks, with a pronounced tendency to commit more Type II errors (False Negatives) than Type I errors (False Positives) across all experimental setups and task conditions. This finding suggests a stronger tendency toward under- rather than over-estimation. This trend is evident in OD but is less pronounced in segmentation, possibly due to the differing cognitive demands of the two tasks. Annotators' accuracy in correctly identifying objects is higher in a setup involving majority voting (independent process) rather than setup integrating double-check review (dependent process) as well as in group setups compared with individual setups, and in tasks where objects are closely grouped together (dense-target task), as opposed to tasks where objects are dispersed over a wide area (sparse-target task). However, there is a minor difference in the \textit{complete} identification of all actual objects across various setups and task conditions, indicating a similar proportion of Type II errors. Additionally, in our case, experts were not found to improve the quality of annotations, and assigning double weight to annotators with higher expertise does not enhance performance in either correct identification or complete detection of actual objects. In segmentation tasks, weighting expert contributions even increased the number of mapped pixels, leading to an overestimation of the target boundary.

\section*{Acknowledgment}
We thank The Israel Science Foundation (ISF), grant number 299/23, for supporting the project. We thank the annotators who participated in the project and the ethics committee for reviewing and approving the experiment.
Special thanks are extended to Oded Rotem from the Department of Software and Information Systems Engineering, whose insightful analysis of the YOLO model's errors inspired us to examine the performance of human annotators.

\bibliographystyle{IEEEtran}
\bibliography{references}

\vspace{-5em}
\begin{IEEEbiography}[{\includegraphics[width=1in,height=1.25in,clip,keepaspectratio]{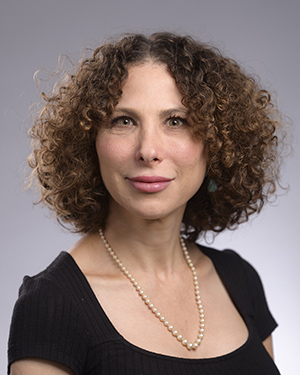}}]{Roni Blushtein-Livnon}
is a Ph.D. student in the Department of Environmental, Geoinformatics, and Urban Planning Sciences at Ben-Gurion University of the Negev. Her doctoral research involves, among other things, Computer Vision and focuses on deep learning methods for object detection and mapping.
\end{IEEEbiography}
\vspace{-5em}
\begin{IEEEbiography}[{\includegraphics[width=1in,height=1.25in,clip,keepaspectratio]{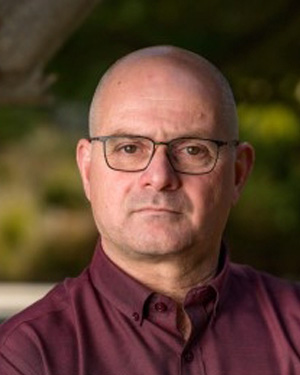}}]{Tal Svoray}
received a Ph.D. in Radar Remote Sensing of Mediterranean Vegetation from Bar Ilan University, Ramat-Gan, Israel, in 2001. 
He is currently a Professor at the Ben-Gurion University of the Negev. His main research interests include object segmentation and detection, remote sensing of soil and vegetation, environmental psychology and geostatistics.
\end{IEEEbiography}
\vspace{-5em}
\begin{IEEEbiography}[{\includegraphics[width=1in,height=1.25in, clip,keepaspectratio]{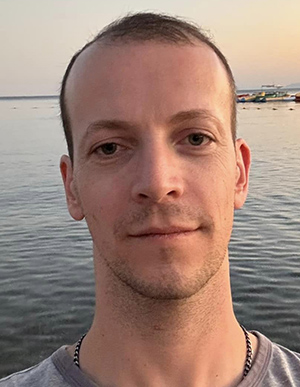}}]{Michael Dorman,} Ph.D., is a programmer and lecturer at The Department of Environmental, Geoinformatics and Urban Planning Sciences, Ben-Gurion University of the Negev. He is working with researchers and students to develop computational workflows for spatial analysis, mostly through programming in Python, R, and JavaScript, as well as teaching those participants. 

\end{IEEEbiography}
\end{document}